# An Attack Traffic Identification Method Based on Temporal Spectrum

Wenwei Xie[1] (IEEE Member); Jie Yin[2]*; Zihao Chen[3]

(1. Trend Micro Nanjing Branch, Nanjing 210012, Jiangsu, China)
(2. Jiangsu Police Institute (JSPI), Nanjing 210031, Jiangsu, China)
(3. University of California, Los Angeles (UCLA), Los Angeles, 90049, CA, USA)

**Abstract:** To address the issues of insufficient robustness, unstable features, and data noise interference in existing network attack detection and identification models, this paper proposes an attack traffic detection and identification method based on temporal spectrum. First, traffic data is segmented by a sliding window to construct a feature sequence and a corresponding label sequence for network traffic. Next, the proposed spectral label generation methods, SSPE and COAP, are applied to transform the label sequence into spectral labels and the feature sequence into temporal features. Spectral labels and temporal features are used to capture and represent behavioral patterns of attacks. Finally, the constructed temporal features and spectral labels are used to train models, which subsequently detects and identifies network attack behaviors. Experimental results demonstrate that compared to traditional methods, models trained with the SSPE or COAP method improve identification accuracy by 10%, and exhibit strong robustness, particularly in noisy environments.

**Keywords:** Machine Learning; Network Security; IPS; Anomaly Detection; Spectrum Label Construction.

　　With the increasing complexity of network environments and the continuous evolution of attack techniques, identifying network attack traffic has become a significant challenge. Traditional detection methods[1][2] struggle to effectively extract and capture dynamic features when dealing with complex network traffic, resulting in reduced model generalization capabilities. Additionally, the pervasive noise in network traffic often undermines the robustness of these models in real-world applications, leading to unstable detection results. Faced with emerging threats and ever-changing attack methods, traditional approaches lack adaptability, limiting their applicability in dynamic network environments.

In recent years, machine learning-based methods for network attack traffic detection[3][4] have garnered significant attention due to their advantages in extracting complex features. However, these methods exhibit notable limitations in handling dynamic traffic characteristics and unstable data environments, primarily in the following two aspects:

1. Limited capability in capturing temporal features: Although deep learning models demonstrate exceptional feature extraction abilities, they face challenges in processing the temporal characteristics of network traffic. Particularly, with long time-series data, capturing both global and local temporal dependencies remains difficult, it is one reason which downgrades model's detection accuracy.

2. Insufficient model robustness: Current models often downgrade performance when confronted with noisy data, they are easily influenced by random noise, which reduces detection accuracy. This issue primarily arises from unstable features and inadequate handling of random noise.

To address these challenges, this paper proposes a malicious traffic identification method based on temporal-spectrum analysis. To address the issue of insufficient temporal feature extraction, a sliding window is employed to segment network traffic, constructing temporal features that retain the fundamental characteristics of traffic while capturing its temporal information. To enhance model robustness, the SSPE/COAP algorithm is introduced to construct spectrum labels. By converting original discrete binary labels into continuous spectrum labels, the model's noise resistance is improved, thereby enhancing detection robustness. The main contributions of this paper are as follows:

1. Proposed the SSPE and COAP algorithms to transform discrete labels into continuous spectrum labels, improving label representation.

2. Utilized a sliding window to process raw traffic data, constructing temporal features that preserve temporal information and enhance feature representation.

3. Conducted experiments on the Edge-IIoTset dataset[5], demonstrating that models trained with SSPE/COAP not only improve detection accuracy but also exhibit superior noise resistance.

## 1  1 Literature Review

In early research, Gupta et al.[6] proposed a network intrusion prevention system (IPS) model based on tree classifiers, focusing on anomaly detection in IoT traffic. The model's effectiveness was validated within IoT environments. Ma et al. [7] combined support vector machines (SVMs) with clustering methods to classify traffic features, this method significantly enhances detection capabilities in complex network environments. Traditional machine learning methods typically select features with high relevance with target variables form an optimal feature subset to reduce data dimensionality and computational costs. These features are then used to train classifiers, which ultimately detect and identify attack traffic.

However, traditional machine learning approaches often focus only on specific key features of network traffic, neglecting inter-feature relationships. Due to the massive scale and complex structure of network traffic data, these methods face limitations in processing capabilities, making it difficult to perform comprehensive and accurate analysis, which results in frequent false positives and false negatives.

As attack types grow increasingly complex, traditional methods show significant shortcomings in handling high-dimensional data and diverse attack patterns. Their performance gradually weakens in emerging, complex network environments, making reliable detection increasingly difficult[8].

Deep learning methods have recently become the mainstream for network traffic analysis. Wang et al.[9] introduced a model combining Generative Adversarial Networks (GANs) and Transformer architectures, achieving improved accuracy in identifying malicious behaviors in complex network traffic. Jung et al. [10] employed Autoencoders to capture both continuous and discontinuous patterns in data, detecting anomalies through reconstruction errors, effectively handling large-scale network traffic. Zamanzadeh et al. [11] highlighted the ability of deep learning to capture temporal and spatial features, significantly improving the accuracy and robustness of time-series anomaly detection.

Furthermore, Feng et al.[12] utilized Convolutional Neural Networks (CNNs) to classify VPN traffic, addressing inefficiencies in traditional methods when dealing with encrypted and obfuscated traffic. CNN-based feature extraction from raw data enhanced classification accuracy, particularly for encrypted traffic, achieving higher identification rates in real-world network environments. Long Short-Term Memory (LSTM) networks, well-suited for handling sequential data, excel in capturing temporal characteristics such as timestamps and durations in traffic data[13], Bahe et al. [14] demonstrated the superiority of LSTM in traffic aggregation prediction, showing its ability to uncover intrinsic correlations among features.

Existing studies reveal that both traditional and deep learning methods have unique strengths in network traffic attack detection. Traditional approaches like SVM and clustering techniques achieved remarkable success in the early stages but struggle with high-dimensional and complex traffic[8]. Although deep learning excels in feature extraction and temporal modeling, its dependency on large, high-quality datasets makes models vulnerable to instability when processing noisy data[15][16].

To enhance model robustness, techniques such as adversarial training and data augmentation can be applied[17]. However, network traffic often suffers from imbalanced sample distributions, where normal samples far outnumber abnormal ones. Constructing balanced datasets demands substantial time and effort. Additionally, deep learning methods require significant computational resources, making it challenging to meet the demands of real-time analysis.

Deep learning techniques demonstrate substantial advantages in capturing complex features and temporal dependencies. To better extract time-series features, Tang et al. [18] proposed a multi-network anomaly detection framework combining label neural networks with LSTM. By learning business logic labels, this approach improved model robustness and detection accuracy. Sørbø et al.[19] explored the impact of label generation strategies on detection performance, emphasizing the importance of label generation methods in enhancing detection outcomes. Pekar et al.[20] introduced a network attack detection method based on advanced flow label generation. Through improved data flow labeling and spectral analysis, their model exhibited enhanced adaptability to dynamic traffic in real-world scenarios.

## 2 Research Methodology

### 2.1 Overall Framework

The overall framework, as illustrated in Figure 1, consists of two main components: Dataset Construction and Model Training. The "Dataset Construction" component involves converting raw network packets and labels into datasets for model training, while the "Model Training" component focuses on training regression or classification models and evaluating their performance.

**Dataset Construction:** First, network data is extracted from PCAP files and segmented into multiple traffic sequences and corresponding label sequences with a sliding window. Each traffic sequence is a set of sequential network packets, and each label sequence is a set of original labels. Next, through feature selection and a flattening operation, the two-dimensional traffic sequences are transformed into traffic features, and the SSPE/COAP algorithm is applied to convert label sequences into spectrum-based labels. Finally, the constructed traffic features and spectrum labels are used to create a new dataset for subsequent model training.

**Model Training:** This study employs two sets of experiments to validate the effectiveness of SSPE/COAP in detecting and identifying attack traffic by training classification and regression models: The dataset undergoes standardization, sampling, and splitting into training and testing sets before training models. For the attack detection task, the spectrum labels are binarized and used as training labels to train a binary classification model, with its performance evaluated. For the attack identification task, the spectrum labels are directly used as training labels, and the model outputs are analyzed to identify attack types based on the distribution of spectrum values.

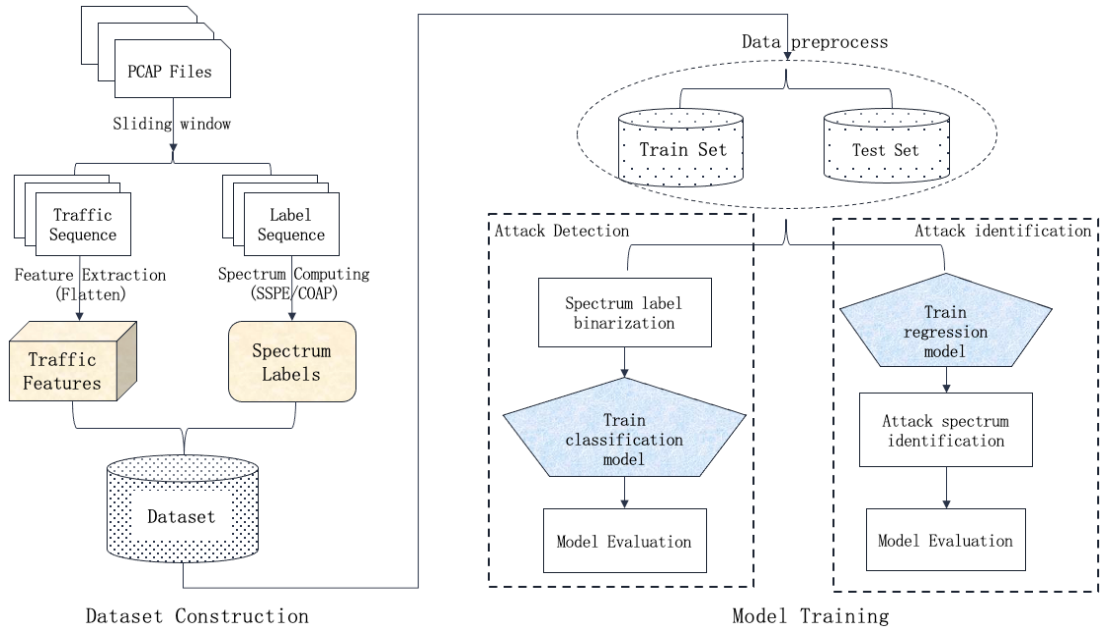

Figure 1: Overall Framework Diagram

## 2.2 Traffic Feature Construction

Traffic feature construction involves four main steps: Missing Temporal Point Filling, Feature Selection, Traffic Sequence and Label Sequence Generation, and Feature Aggregation.

**Missing Temporal Point Filling:** In raw traffic data, missing temporal points are common. To address this issue, this study randomly samples normal traffic data from normal samples to fill the missing points, ensuring the continuity of the time series. This approach facilitates better feature extraction. Figure 2 illustrates the results of the filling process.

| ID | time | proto | ··· | size | label |
|---|---|---|---|---|---|
| 1 | 16:48:47 | tcp | ··· | 74 | 1 |
| 2 | 16:48:48 | ssdp | ··· | 208 | 0 |
| 3 | 16:48:49 | ssdp | ··· | 208 | 0 |
| 4 | 16:48:50 | ssdp | ··· | 208 | 0 |
| 5 | 16:48:51 | ssdp | ··· | 208 | 0 |
| 6 | 16:48:52 | ip:tcp | ··· | 66 | 0 |
| 7 | | | | | |
| 8 | | | | | |
| 9 | 16:48:55 | igmp | ··· | 46 | 0 |
| 10 | | | | | |
| 11 | | | | | |
| 12 | | | | | |
| 13 | 16:48:59 | igmp | ··· | 46 | 0 |
| 14 | 16:49:00 | tcp | ··· | 74 | 1 |
| ··· | ··· | ··· | ··· | ··· | ··· |

网络流量（填充前）

| ID | time | proto | ··· | size | label |
|---|---|---|---|---|---|
| 1 | 16:48:47 | tcp | ··· | 74 | 1 |
| 2 | 16:48:48 | ssdp | ··· | 208 | 0 |
| 3 | 16:48:49 | ssdp | ··· | 208 | 0 |
| 4 | 16:48:50 | ssdp | ··· | 208 | 0 |
| 5 | 16:48:51 | ssdp | ··· | 208 | 0 |
| 6 | 16:48:52 | tcp | ··· | 66 | 0 |
| 7 | 16:48:53 | tcp | ··· | 54 | 0 |
| 8 | 16:48:54 | mqtt | ··· | 56 | 0 |
| 9 | 16:48:55 | igmp | ··· | 46 | 0 |
| 10 | 16:48:56 | tcp | ··· | 54 | 0 |
| 11 | 16:48:57 | dns | ··· | 81 | 0 |
| 12 | 16:48:58 | tcp | ··· | 54 | 0 |
| 13 | 16:48:59 | igmp | ··· | 46 | 0 |
| 14 | 16:49:00 | tcp | ··· | 74 | 1 |
| ··· | ··· | ··· | ··· | ··· | ··· |

网络流量（填充后）

Figure 2: Missing Temporal Point Filling

**Feature Selection:** Initially, manual feature selection methods are applied to remove irrelevant content from the raw data packets. Features specific to different attack types are then selected. Table 1 shows the number of features chosen for each attack type.

Table 1 Selected feature count

| Attack | Features | Attack | Features |
|---|---|---|---|
| Port Scanning attack | 37 | Ransomware attack | 40 |
| Vulnerability scanner attack | 56 | DDoS HTTP Flood Attacks | 48 |
| Password attacks | 48 | Backdoor_attack | 37 |
| Uploading attack | 40 | MITM (ARP spoofing + DNS) Attack | 23 |
| DDoS UDP Flood Attacks | 9 | SQL injection attack | 37 |
| DDoS ICMP Flood Attacks | 15 | XSS attacks | 48 |
| DDoS TCP SYN Flood Attacks | 32 | OS Fingerprinting attack | 35 |

**Traffic Sequence and Label Sequence Generation:** To preserve the temporal information of network traffic, a sliding window is used to segment the traffic into multiple traffic sequences and label sequences. As the sliding window moves, the features within the window form a traffic sequence, while the labels within the window form a label sequence. This approach retains the temporal characteristics of the traffic, enabling the model to better capture the sequential features of attack behaviors. Figure 3 illustrates the process of generating traffic sequences and label sequences.

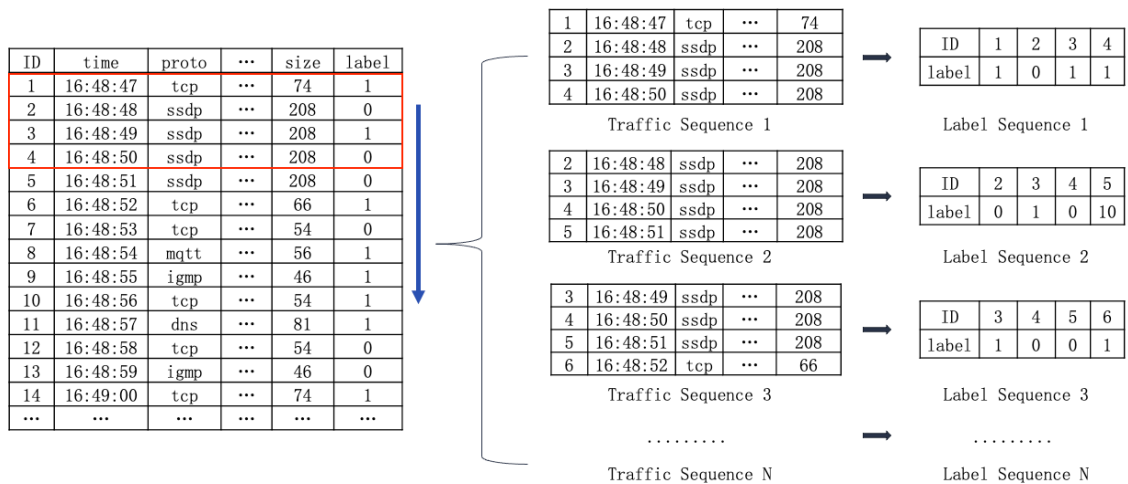

Figure 3: Traffic Sequence and Label Sequence Generation

**Feature Aggregation:** The Flatten Aggregation method is employed to flatten the two-dimensional traffic sequences within the sliding window into one-dimensional feature vectors. These traffic features retain all information from the traffic sequences, including protocol types, packet sizes, and more. Figure 4 illustrates the feature aggregation process.

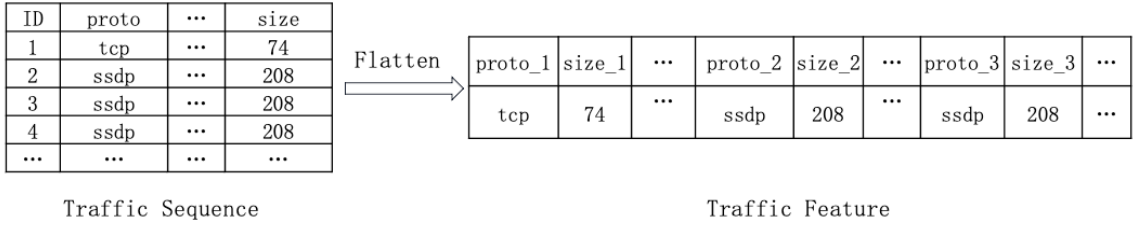

Figure 4: Feature Aggregation

2.3 Spectrum Label Construction

The label sequence is a binary time series, where 0 represents normal traffic and 1 represents attack traffic. Spectrum label construction maps the label sequence to a real number, which serves as the spectrum label. This study employs two methods for generating spectrum labels: COAP (Count of Attack Packets) and SSPE (Sum of Sinusoidal Positional Encoding).

**COAP Method:** The COAP method disregards positional information within the sequence and directly counts the number of attack packets. COAP intuitively reflects the intensity of an attack, making it suitable for scenarios requiring rapid evaluation of attack frequency. The calculation is defined as shown in Formula 1:

$$L_{COAP} = \sum_{pos=0}^{N} Label_{pos} \qquad (1)$$

Where:

- $N$ is the length of the label sequence.
- $Label_{pos} \in \{0,1\}$ is the original label of the $pos$ number packet.
  - A label of 0 indicates normal traffic.
  - A label of 1 indicates abnormal (attack) traffic.
- $L_{COAP}$ is the spectrum label.

**SSPE Method:** To address the loss of positional information in the COAP method, this study adopts the sinusoidal positional encoding approach as referenced in the paper[21]. This method encodes the positions in the traffic sequence using sine and cosine functions. The position encodings are then multiplied with the corresponding label values and summed to generate the spectrum label.

Since the label sequence is a binary time series, multiplication can effectively filter out the influence of normal traffic (with label value 0), leaving only the information related to attack traffic. SSPE can more precisely describe the temporal changes in attack behavior. The calculation method is shown in Formula 2.

$$L_{SSPE} = \sum_{pos=0}^{N} \sum_{i=0}^{d_{model}} Label_{pos} \times PE_{pos,i} \qquad (2)$$

Where:

- $N$ is the length of the label sequence.
- $Label_{pos} \in \{0,1\}$ is the original label of the $pos$ number packet.
    - A label of 0 indicates normal traffic.
    - A label of 1 indicates abnormal (attack) traffic.
- $d_{model}$ is a hyperparameter that represents the encoding dimension.
- $PE_{pos,i}$ is the position encode of the $pos$ and the $i$ dimension.
- $L_{SSPE}$ is the spectral label.

$PE_{pos}$ represents the position encode of the $pos$ packet, and the calculation method is shown in Formula 3.

$$PE_{pos} \begin{cases} PE_{(pos,2i)} = \sin(\frac{pos}{10000^{\frac{2i}{d_{model}}}}) \\ PE_{(pos,2i+1)} = \cos(\frac{pos}{10000^{\frac{2i}{d_{model}}}}) \end{cases} \qquad (3)$$

Where:

- $d_{model}$ is a hyperparameter that represents the encoding dimension.
- $i$ is the dimension index.

2.4 Data Preprocessing

The Z-Score method is used to standardize the feature values, with the calculation method shown in Formula 4.

$$x'_{i,j} = \frac{x_{i,j} - \mu_j}{\sigma_j} \qquad (4)$$

Where:

- $x_{i,j}$ is the value of the $j$ feature of the $i$ sample.
- $\mu_j$ is the mean of the $j$ feature.
- $\sigma_j$ is the deviation of the $j$ feature.
- $x'_{i,j}$ is the standardized value the $j$ feature for the $i$ sample.

2.5 Model Training

This study verifies the model's robustness with noise through two tasks: attack detection and attack identification. The attack detection task uses a binary classification model to distinguish the normal and attack traffic, without identifying the type of attack traffic. The attack identification task uses a regression model to output the spectrum labels of the traffic, and the attack type is identified based on the similarity of the spectrum labels distribution.

**Spectrum label binarization:** The constructed spectrum labels are continuous values. Before training the binary classification model, the spectrum labels are binarized with a threshold. The processing method is shown in Formula 5

$$x' = \begin{cases} 1 & if\ x \geq \tau \\ 0 & if\ x < \tau \end{cases} \tag{5}$$

Where:

- $x$ is the spectrum label.
- $\tau$ is the pre-defined threshold.
- $x'$ is the binarized label.

In this study, the threshold $\tau$ is determined based on the percentile corresponding to the proportion of attack samples. The calculation method is shown in Formula 6.

$$\tau = Percentile(X, \frac{N_1}{N} \times 100) \tag{6}$$

Where:

- $X$ is the spectrum label sequence arranged in ascending order.
- $N$ is the length of the sequence.
- $N_1$ is the number of elements in the sequence with a label of 1.

- $\tau$ is the selected threshold.

**Attack spectrum identification:** The attack type is identified by calculating the similarity between the model's output spectrum labels distribution and the true spectrum labels distribution. Cosine similarity[22] is used for the similarity calculation. The method is shown in Formula 7.

$$c^* = argmin_i(1 - \frac{y \cdot x_i}{\|y\|\|x_i\|}) \qquad (7)$$

Where:

- $y$ is the spectrum labels distribution vector output by the model.
- $x_i$ is the spectrum labels distribution vector of the $i$ attack in the dataset.
- $c^*$ is the identified attack type.

**Model Selection:** In this study, AutoML tool H2O (V3.46.0.4) was employed for training. Total five models, including ANN, GBM, GLM, RF, and XGBoost, were selected for classification and regression tasks. The modified hyperparameters are shown in Table 2 (with default values for unchanged parameters).

Table 2: Modified Model Hyperparameters

| Model Type | Parameters (Classification Model) | Parameters (Regression Model) |
|---|---|---|
| ANN(Artificial Neural Network) | Hidden=[100, 100] | Hidden=[100, 100] |
| GBM(Gradient Boosting Machine) | Ntrees=50 | Ntrees=50 |
| GLM(Generalized Linear Model) | Family=binomial | Family=gaussian |
| RF(Random Forest) | Ntrees=25, max_depth=10 | Ntrees=25, max depth=10 |
| XGBoost(Extreme Gradient Boosting) | n_estimators=100 | n_estimators=100 |

2.6 Model Evaluation

In the attack detection task, the binary classification model directly outputs classification results, and is evaluated using common classification metrics, including Accuracy, Precision, Recall, and F1-score.

In the attack identification task, the identification is performed based on similarity, and the evaluation metric is the identification accuracy, with the calculation method shown in Formula 8.

$$Accuracy = \frac{Correctly\ Identified\ Samples}{Total\ Samples} \qquad (8)$$

**3 Simulation Experiments**

3.1 Dataset

In this study, the dataset Edge-IIoTset[5] was used, which is a network security dataset specifically designed for the Industrial Internet of Things (IIoT). The dataset was generated by capturing network packets with the tool of tcpdump to create PCAP files, and then using tools such as Zeek and TShark for session aggregation and feature extraction. The Edge-IIoTset dataset contains 14 attack types related to IoT and IIoT devices, which are categorized into five major attack types: Denial of Service (DoS/DDoS) attacks, information gathering, man-in-the-middle attacks, injection attacks, and malware attacks.

3.2 Experimental Methods and Benchmarks

This study builds upon the experiment in of the dataset Edge-IIoTset by simulating real-world application scenarios through the addition of random noise to feature values. The robustness of models trained using the SSPE/COAP methods was evaluated under different noise ratio environments.

**Noise Environment Construction:** First, samples to be noised were randomly selected based on a specified ratio. Then, random noise was added to the feature values of the selected samples. In this study, 11 test sets with varying noise ratios ranging from 0% to 100% (in increments of 10%) were constructed through this process, as detailed in Table 3.

Table 3: Test sets

| Test set | Description | Test set | Description |
|---|---|---|---|
| 1 | No noise | 7 | 60% samples with noise |
| 2 | 10% samples with noise | 8 | 70% samples with noise |
| 3 | 20% samples with noise | 9 | 80% samples with noise |
| 4 | 30% samples with noise | 10 | 90% samples with noise |
| 5 | 40% samples with noise | 11 | 100% samples with noise |
| 6 | 50% samples with noise | | |

## 3.3 Experimental Results

**Spectrum Label Distribution:** During the construction of spectrum labels with SSPE and COAP, the impact of sliding window size and sinusoidal positional encoding dimension on the results was carefully analyzed. Multiple experiments were conducted with varying sliding window sizes (10, 20, 30, 40, 50, 60) and sinusoidal encoding dimensions (2, 4, 8, 16, 32, 64, 128, 236, 256). The optimal sliding window size and sinusoidal encoding dimension were determined based on the similarity and normality of the generated spectrum labels distributions. The selected parameters are shown in Table 4, and the resulting spectrum labels distributions are depicted in Figure 5. The results indicate that compared to COAP, spectrum labels generated using SSPE exhibit more details and are closer to one or multiple normal distributions.

Table 4: PE Dimension and Sliding Window Size

| Attack | PE Dimension | Sliding Window Size |
|---|---|---|
| Vulnerability scanner attack | 8 | 30 |
| Password attacks | 236 | 30 |
| Uploading attack | 8 | 60 |
| DDoS UDP Flood Attacks | 256 | 30 |
| DDoS ICMP Flood Attacks | 8 | 30 |
| Ransomware attack | 128 | 30 |
| Backdoor_attack | 256 | 60 |
| MITM (ARP spoofing + DNS) Attack | 128 | 60 |
| SQL injection attack | 128 | 60 |
| OS Fingerprinting attack | 64 | 60 |
| Port Scanning attack | 128 | 60 |
| DDoS TCP SYN Flood Attacks | 64 | 60 |
| DDoS HTTP Flood Attacks | 16 | 60 |
| XSS attacks | 256 | 60 |

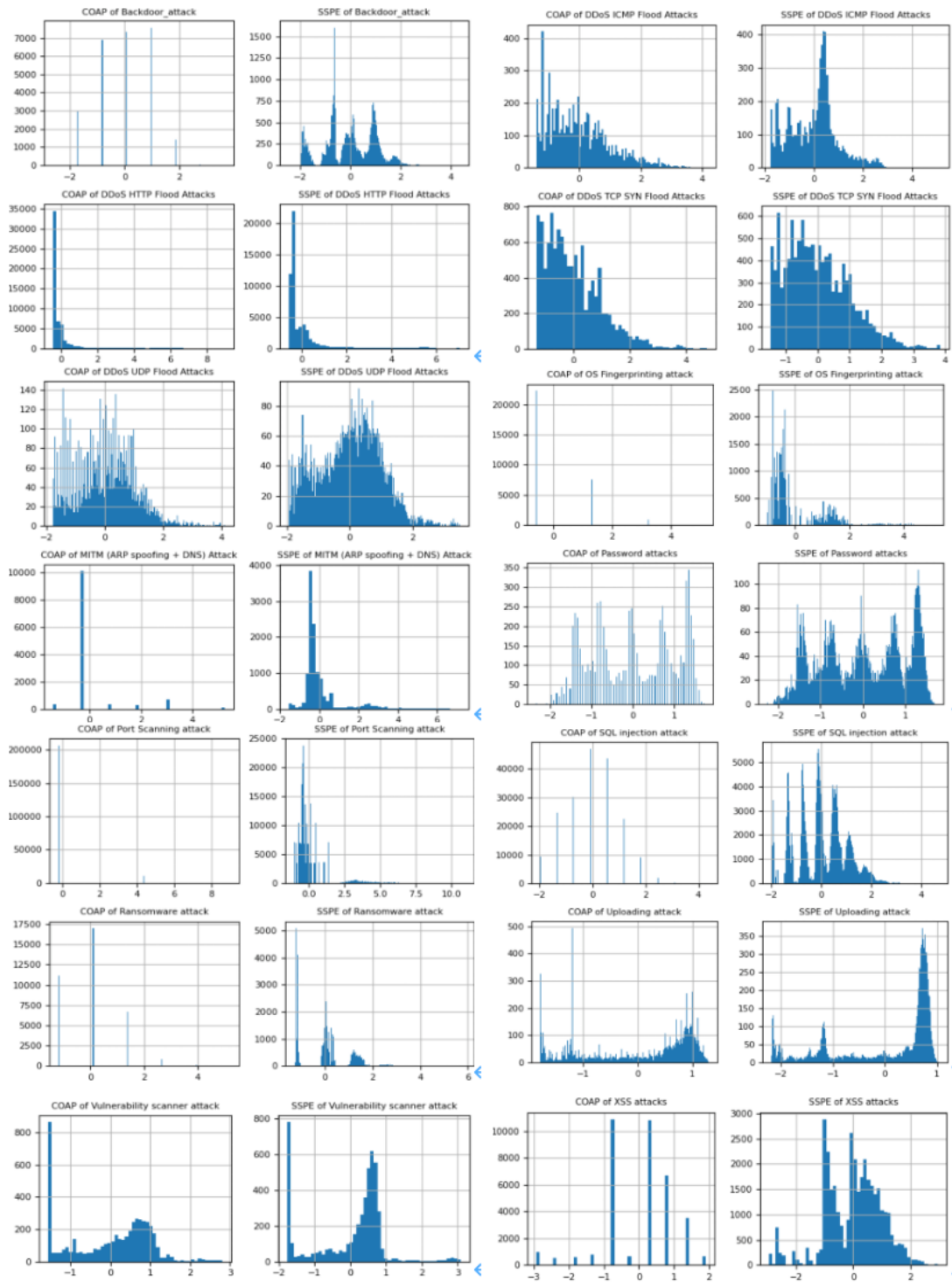

Figure 5: Histograms of Spectrum Labels Generated by COAP and SSPE

**Evaluation of Attack Detection Task:** In the attack detection task, we trained five classification models: ANN, GBM, GLM, RF, and XGBoost. These models were tested in environments with varying noise ratios, and their performance was evaluated by metrics including Recall, Precision, Accuracy, and F1-score.

The average performance of these models with noise ratio is shown in Figure 6 (with the horizontal axis is the noise ratio and the vertical axis is the evaluation metrics). In an environment with a noise ratio of 100% (all samples are noised), the performance of each model is detailed in Table 5.

The results show that models trained using traditional methods experience a rapid performance decline as the noise ratio increases. In contrast, models trained using the SSPE or COAP methods demonstrate significantly better noise resistance, with comparable robustness between the SSPE and COAP methods.

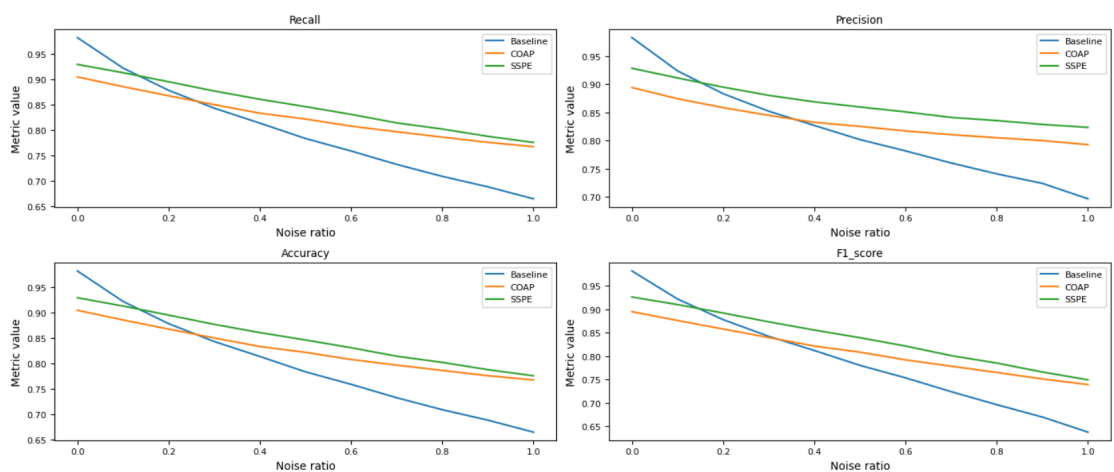

Figure 6: Average performance with noise in attack detection task

Table 5: Model performance tested with 100% noised samples.

| Model | Method | Accuracy | Recall | Precision | F1-score |
|---|---|---|---|---|---|
| ANN | Baseline | 0.72292029 | 0.72292029 | 0.72296494 | 0.72286184 |
| | COAP | 0.77565133 | 0.77565133 | 0.81730491 | 0.75735556 |
| | SSPE | 0.71006356 | 0.71006356 | 0.79616271 | 0.6664234 |
| GBM | Baseline | 0.63077518 | 0.63077518 | 0.65016755 | 0.61982796 |
| | COAP | 0.69222755 | 0.69222755 | 0.66198181 | 0.62805779 |
| | SSPE | 0.74091954 | 0.74091954 | 0.78158745 | 0.69910985 |
| GLM | Baseline | 0.7547952 | 0.7547952 | 0.75654822 | 0.75450875 |
| | COAP | 0.7436931 | 0.7436931 | 0.79551599 | 0.72068684 |
| | SSPE | 0.77310653 | 0.77310653 | 0.81989329 | 0.74864341 |
| RF | Baseline | 0.69162532 | 0.69162532 | 0.70265787 | 0.68794904 |
| | COAP | 0.8069137 | 0.8069137 | 0.83000332 | 0.78604388 |
| | SSPE | 0.85702648 | 0.85702648 | 0.88207817 | 0.84961954 |
| XGBoost | Baseline | 0.5243465 | 0.5243465 | 0.65092504 | 0.40438943 |
| | COAP | 0.81896748 | 0.81896748 | 0.85854683 | 0.80460315 |
| | SSPE | 0.79753268 | 0.79753268 | 0.83681568 | 0.78372926 |

In a fully noisy environment (where 100% of the samples contain random noise), models trained using the SSPE and COAP methods achieved an average accuracy improvement of 10% compared to those trained with the baseline method. As presented in Table 6.

Table 6: Model average performance improvement

| Method | Accuracy |
| --- | --- |
| Baseline | 0.664892497 |
| COAP | 0.767490632 +10.26% |
| SSPE | 0.775729759 +11.08% |

**Evaluation of Attack Identification Task:** We trained five regression models: ANN, GBM, GLM, RF, and XGBoost, to identify attack types by calculating the similarity between the model output spectrum labels distribution and the true spectrum labels distribution. These models were tested in environments with varying noise ratios to evaluate their accuracy. The average accuracy of these five models as a function of noise ratio is shown in Figure 7 (the horizontal axis is the noise ratio, and the vertical axis is identification accuracy). In an environment with a noise ratio of 100% (where all samples are noised), the performance of each model is shown in the table 7. The results indicate that models trained using SSPE perform better in the attack identification task.

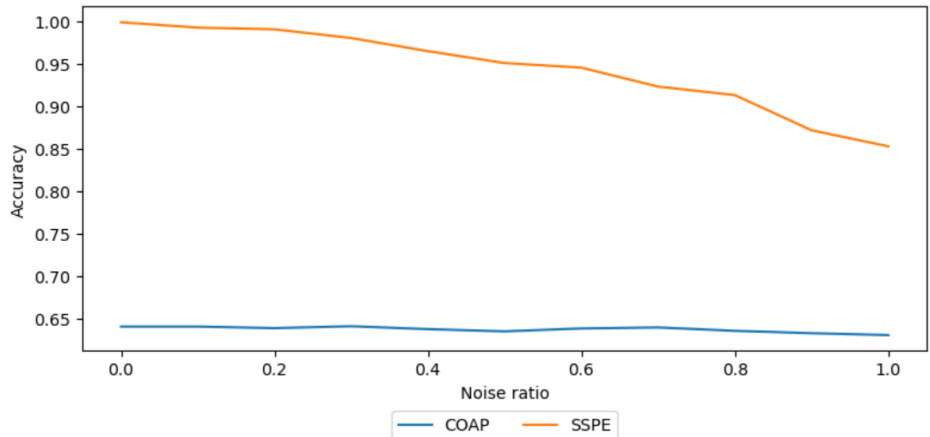

Figure 7: Average model identification accuracy with noise

Table 7: Average model identification accuracy

| Model | COAP | SSPE |
|---|---|---|
| ANN | 0.615 | 0.87857143 |
| GBM | 0.627 | 0.78571429 |
| GLM | 0.649 | 0.93 |
| RF | 0.621 | 0.83 |
| XGBoost | 0.64 | 0.83857143 |

## 3. Conclusion and Future Work

This paper presents a method for identifying attack traffic through time-series spectra labels, aiming to address issues such as insufficient model robustness, unstable features, and noise interference in existing network traffic attack detection and identification tasks. The method segments the traffic using a sliding window, applies the Flatten operation to convert traffic sequences into features, and converts label sequences into spectrum labels using the SSPE or COAP algorithms.

Experimental results show that models trained on datasets constructed using SSPE or COAP outperform traditional machine learning methods in both attack detection and identification tasks, particularly in noisy environments, where the models exhibit stronger robustness. In a fully noisy environment, SSPE and COAP methods achieved 10% improvement of average accuracy.

In the future, we will deeply research in the following two areas: First, we will explore combining spectrum generation methods with deep neural networks and large models to enhance the model's expressive ability and improve network attack identification accuracy. Second, we will experiment with SSPE and COAP algorithms to transform feature values, enriching the features to further enhance the network attack recognition capability.

Author Biographies:

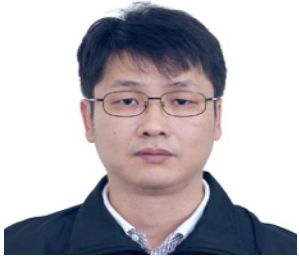

**Wenwei Xie**: IEEE Member, received a master's degree in MBA from the Nanjing University of Aeronautics and Astronautics (NUAA), Nanjing, China in 2013. He is currently working toward network security in Trend Micro Incorporated, Nanjing, China. His research interests include Cyber security, AI technology, and Computer vision.
His e-mail is jim.xie.cn@outlook.com.

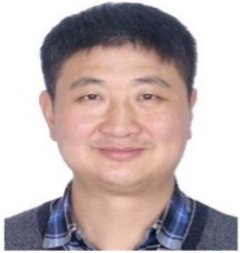

**Jie Yin:** received M.S. degree in Software Engineering from Nanjing University of Science and Technology, Nanjing, China in 2008. He is currently a Senior Engineering with the Department of Computer Information and Cyber Security, Jiangsu Police Institute, Nanjing, China. His recent research interests include machine learning, big data, and network security.
His e-mail is yinjie@jspi.cn.

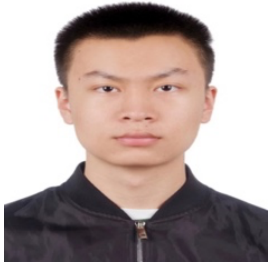

**Zihao Chen:** received the bachelor's degree in Statistics from University of Illinois, Urbana-Champaign（UIUC）, United States in 2023. He is currently studying Data Science and Machine Learning in University of California Los Angeles, Westwood, United States. His research interests include Data Engineering, Neural Network and Statistical Theory.
His e-mail is czh0711@g.ucla.edu.